# Zero-Shot Generative De-identification: Inversion-Free Flow for Privacy-Preserving Skin Image Analysis

Konstantinos Moutselos*[1] [0000-0002-6759-8540] and Ilias Maglogiannis[1] [0000-0003-2860-399X]

[1] University of Piraeus, Dept. of Digital Systems. Piraeus 18534, Greece
{kmouts;imaglo}@unipi.gr

## Abstract

*The secure analysis of dermatological images in clinical environments is fundamentally restricted by the critical trade-off between patient privacy and the preservation of diagnostic fidelity. Traditional de-identification techniques often degrade essential pathological markers, while state-of-the-art generative approaches typically require computationally intensive inversion processes or extensive task-specific fine-tuning, limiting their feasibility for real-time deployment.*

*This study introduces a zero-shot generative de-identification framework that utilizes an inversion-free pipeline for privacy-preserving medical image analysis. By leveraging Rectified Flow Transformers (FlowEdit), the proposed method achieves high-fidelity identity transformation in less than 20 seconds without requiring pathology-specific training or labeled datasets. We introduce a novel "segment-by-synthesis" mechanism that generates counterfactual "healthy" and "pathological" digital twin pairs to isolate clinical signals from biometric identifiers in a zero-shot manner. Our approach specifically utilizes the CIELAB color space to decouple erythema-related pathological signals from semantic noise and individual skin characteristics.*

*Pilot validation on high-resolution clinical samples demonstrates robust stability in preserving pathological features, achieving an Intersection over Union (IoU) stability exceeding 0.67, while ensuring rigorous de-identification. These results suggest that the proposed zero-shot, inversion-free approach provides a scalable and efficient solution for secure data sharing and collaborative biomedical research, bypassing the need for large-scale annotated medical datasets while aligning with data protection standards.*

## 1. Introduction

The clinical analysis of skin images is a cornerstone of modern dermatology, yet its progress is increasingly hindered by the sensitive nature of patient biometric data. High-resolution dermatological photography, while essential for the accurate monitoring of conditions such as erythema, frequently captures identifiable features that pose significant privacy risks. In compliance with stringent data protection standards like the GDPR, robust de-identification is mandatory for the sharing of medical datasets and collaborative research. The integration of Generative AI into dermatological analysis models promises to democratize diagnostic access, building on a growing body of work in automated detection [1] for diseases such as dermatitis and vitiligo [2]. However, widespread adoption is hampered by the ethical imperative of patient privacy. Unlike radiological scans, which can be easily anonymized, facial skin images contain immutable biometric markers—such as ocular structure and craniofacial geometry—that pose severe re-identification risks.

Traditional anonymization techniques, such as blurring or pixelation, often compromise the diagnostic utility of the image by obscuring fine-grained pathological markers. While generative models, particularly those based on Diffusion, have shown promise in synthesizing identity-agnostic 'digital twins', they often rely on computationally intensive inversion processes. Such overhead is impractical for real-time clinical workflows or deployment on resource-constrained edge devices. To address these challenges, we present an inversion-free generative framework for privacy-preserving skin image analysis that prioritizes both identity protection and pathological integrity

### 1.1. Problem: Identity as Semantic Noise and Privacy Risk

Current zero-shot segmentation models, such as Grounded-SAM [3], operate on semantic priors that are ill-suited for the subtle, diffuse nature of skin inflammation. These models create a dual-failure mode in the FL context:

1. Diagnostic Drift: Pathological erythema is frequently conflated with nonclinical distractors, such as piercings, lips, or natural skin flushing, introducing “noisy labels” into the federated model.
2. Biometric Leakage: The segmentation mask inadvertently encodes identity-specific contours (e.g., the exact shape of an eye or a unique tattoo) by failing to isolate the pathology, exposing the local client to gradient reconstruction attacks [4].

### 1.2. Proposed Architecture: A Generative Privacy Layer for Clinical Data

We propose a generative framework designed to function as an autonomous privacy layer at the point of clinical data acquisition. Unlike centralized anonymization services that require the transfer of sensitive records, our approach operates locally on clinical edge devices. By utilizing an inversion-free flow matching mechanism, the system acts as a 'privacy gatekeeper' that transforms identifiable patient photography into diagnostic-equivalent synthetic surrogates. This architecture ensures that only identity-agnostic signals are transmitted for downstream analysis, effectively mitigating the risk of re-identification while maintaining the high-frequency textural details required for dermatological assessment.

The proposed approach uses FlowEdit [5], an inversion-free Rectified Flow Transformer to perform high-fidelity identity transformation. Unlike diffusion models that require computationally intensive inversion loops, our flow-based approach solves a direct ODE, enabling fast, low-compute inference suitable for edge deployment. We create a "privacy firewall" at the data source by shifting the patient's facial structure to a synthetic surrogate while anchoring the erythema in the latent space (Fig. 1).

We further refine this with the counterfactual Segment-by-Synthesis mechanism [6]. By locally generating a "healthy twin" of the synthetic patient, we extract a differential mask via pixel-wise subtraction. This mask is robust to semantic noise and is stripped of identity-linked features to ensure that only safe, agnostic features are used for training.

### 1.3. Contributions

Our pilot study, which was validated on high-resolution clinical samples, offers three primary contributions:

1. Inversion-Free Identity Transformation: We leverage Rectified Flow Transformers (FlowEdit) to perform high-fidelity identity shifts in under 20 seconds. This inversion-free approach significantly reduces computational latency, making generative de-identification viable for edge-native clinical environments.
2. Clinical Signal Isolation via Segment-by-Synthesis: We introduce a novel mechanism to generate locally-consistent healthy and pathological counterfactual pairs. This allows for the precise extraction of differential masks that decouple diagnostic signals (e.g., erythema) from biometric identifiers and semantic noise (e.g., jewelry, specular reflections).
3. Signal-Aware Processing in CIELAB Space: Our

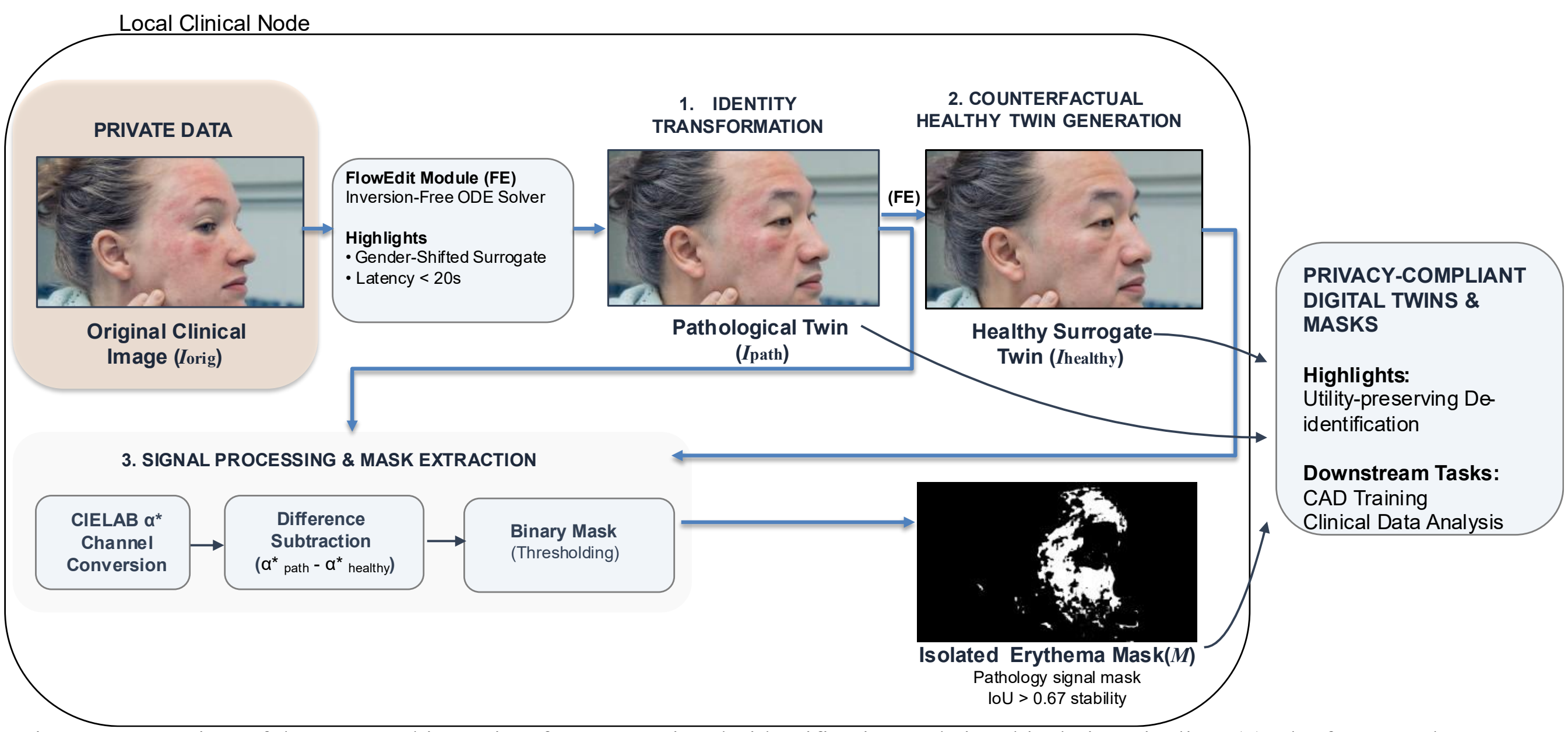

Figure 1: Overview of the proposed inversion-free generative de-identification and signal isolation pipeline. (1) The framework operates at the local clinical node where a raw clinical image ($I_{orig}$) [19] undergoes an identity transformation via the FlowEdit Module (FE). This produces a identity-shifted Pathological Twin ($I_{path}$) that preserves diagnostic markers while scrubbing biometric identifiers. (2) A counterfactual Healthy Surrogate Twin ($I_{healthy}$) is synthesized to serve as a personalized healthy baseline. (3) The clinical signal is isolated using a segment-by-synthesis approach involving CIELAB $\alpha^*$ channel conversion and pixel-wise difference subtraction ($\alpha^*_{path} - \alpha^*_{healthy}$). The resulting isolated Erythema Mask (M) achieves high structural stability (IoU > 0.67), enabling the generation of privacy-compliant digital twins for downstream multi-institutional CAD training and clinical data analysis.

framework incorporates CIELAB color space analysis to ensure the consistent preservation of pathological redness intensity. By isolating clinical signals from pigmentation and lighting artifacts, we maintain the diagnostic utility of the transformed images for subsequent medical analysis.

## 2. Related Work

### 2.1. Privacy Challenges and Re-identification Risks in Clinical Imaging

The sharing of high-resolution dermatological imagery for collaborative research is primarily governed by the need to protect patient biometric privacy [7]. Unlike radiological data, which can often be de-identified through standard metadata removal, facial and skin photography contains unique anatomical markers—including ocular structure and craniofacial geometry—that facilitate unauthorized patient re-identification [8]. In accordance with the General Data Protection Regulation (GDPR) [9], such data is also classified as sensitive Protected Health Information (PHI) [10], necessitating rigorous de-identification prior to multi-institutional analysis. Federated Learning (FL) enables collaborative model training without centralizing patient data [7]. However, recent studies on Gradient Inversion Attacks have demonstrated that patient data can often be reconstructed from shared model updates, particularly in sparse datasets like clinical skin imaging data [4]. Differential Privacy (DP) adds noise to gradients to mitigate this, but it degrades the utility of fine-grained diagnostic features [7]. Our work targets a complementary defense: Input-Level De-identification. We neutralize the risk of reconstruction attacks at the root by scrubbing biometric markers at the client source before the data enters the training loop.

### 2.2. Generative De-identification: From GANs to Inversion-Free Flows

Textural semantics required for dermatological diagnosis are destroyed by traditional anonymization (blurring, pixelation). Generative approaches, such as DeepPrivacy [11], use GANs to swap identities but often suffer from "texture smoothing," rendering them unsuitable for pathological analysis. To edit real images, recent diffusion models [12] require computationally expensive inversion processes [13], which is impractical for resource-constrained edge devices in hospital networks.

Recent advances in Flow Matching [14] and Rectified Flow [15] have addressed this by linearizing the transport map between distributions. FlowEdit [5] leverages this property to perform inversion-free editing. FlowEdit allows for the modification of global identity semantics while anchoring local textures via guidance by computing the difference between source and target velocity fields directly in latent space, a capability we exploit to efficiently preserve erythema at the edge.

### 2.3. Zero-Shot vs. Synthesis-Based Segmentation

In clinical data pipelines, obtaining high-quality expert annotations is often a significant bottleneck. Recent foundation models, such as the Segment Anything Model (SAM) [16] and its high-fidelity successor HQ-SAM [17] have revolutionized general image segmentation. However, even when combined with open-set detectors, such as the Grounded-SAM [3], these models offer zero-shot segmentation but lack domain specificity, frequently confusing pathology with physiological noise. Although specialized preprocessing algorithms have been successful in removing specific artifacts, such as hair [18], they are often limited to single-noise types. In contrast, Segment-by-Synthesis framework [6] functions as a domain-specific, unsupervised annotator. By generating "Digital Twin" labels locally, this approach bypasses the need for manual expert intervention while maintaining high precision. Instead of relying on generic semantic priors, it isolates the clinical signal by comparing synthetic pathological surrogates with their counterfactual healthy counterparts, ensuring that only the relevant diagnostic information is extracted.

## 3. Methodology

The proposed methodology introduces a localized, edge-native preprocessing framework designed to decouple sensitive biometric signals from diagnostic pathological markers at the point of clinical data acquisition. Unlike centralized anonymization workflows that necessitate the transfer of raw records, this pipeline operates within the secure hospital infrastructure to establish a proactive "privacy firewall". The core objective is to transform identifiable patient photography into privacy-compliant synthetic surrogates before any collaborative analysis or gradient computation occurs.

The framework integrates two primary computational components:

a) Inversion-Free Identity Transformation: Utilizing Rectified Flow Transformers (FlowEdit) to perform high-fidelity de-identification with minimal computational latency.
b) Counterfactual Segment-by-Synthesis: An unsupervised mechanism for isolating the pathological signal (e.g., erythema) from semantic noise by generating "healthy twin" counterparts.

### 3.1. Inversion-Free Identity Transformation (FlowEdit)

To locally de-identify patient images, we use FlowEdit [5], based on Rectified Flow Transformers. Unlike

diffusion models that require iterative inversion to find a noise pivot, FlowEdit solves a direct Ordinary Differential Equation (ODE) to map the source distribution (Patient) to a target distribution (Synthetic Surrogate).

The transformation is governed by the following guided velocity field:

$$\hat{v} = v_{uncond} + \gamma_{src}(v_{src} - v_{uncond}) + \gamma_{tgt}\left(v_{tgt} - v_{uncond}\right) \quad (1)$$

This equation represents the mixed guidance mechanism. It calculates the final direction (velocity) that the latent image needs to move at each time step to satisfy both the need to look like the original image (source) and the need to look like the new identity (target). $\hat{v}$ represents the instantaneous velocity vector field that guides the latent pixels from a noised state toward the target edited distribution. It is sent to the ODE solver to generate the de-identified patient. The Unconditional Velocity ($v_{uncond}$) stands for the model's prediction when given an empty text prompt (""). It represents the "natural drift" of the model—the direction it would take to generate a generic, coherent image based on general statistics (lighting, composition) without any specific semantic instruction. The Source Velocity ( $v_{src}$) contains the directions needed to reconstruct the original patient and their pathology. Is the velocity conditioned on the source prompt (e.g., "Photo of a woman with facial erythema").

The subtraction $v_{src} - v_{uncond}$ isolates the source prompt's pure semantic signal. Generic image features ($v_{uncond}$) are removed to leave only the features specific to the patient's original appearance and disease. The Target Velocity ($v_{tgt}$) *is the* velocity conditioned on the target prompt (e.g., "Photo of a man with facial erythema"). It contains the directions needed to hallucinate the new surrogate identity. The subtraction $v_{tgt} - v_{uncond}$ isolates the pure semantic signal of the target prompt (the new gender/identity features). Both $\gamma_{src}$ *and* $\gamma_{tgt}$ are scalar coefficients. The first determines how strongly the model "clings" to the original image details. The Target Guidance Scale ($\gamma_{tgt}$) pushes the latent geometry toward the new facial structure (e.g., changing the jawline and eyes) while the source guidance ($\gamma_{src}$) fights to maintain the skin texture.

By maintaining high Source Guidance ($\gamma_{src}$), we anchor the pathological textures (erythema) while allowing the global facial geometry to shift. This inversion-free architecture reduces the computational overhead, making it viable for deployment on hospital-grade GPUs.

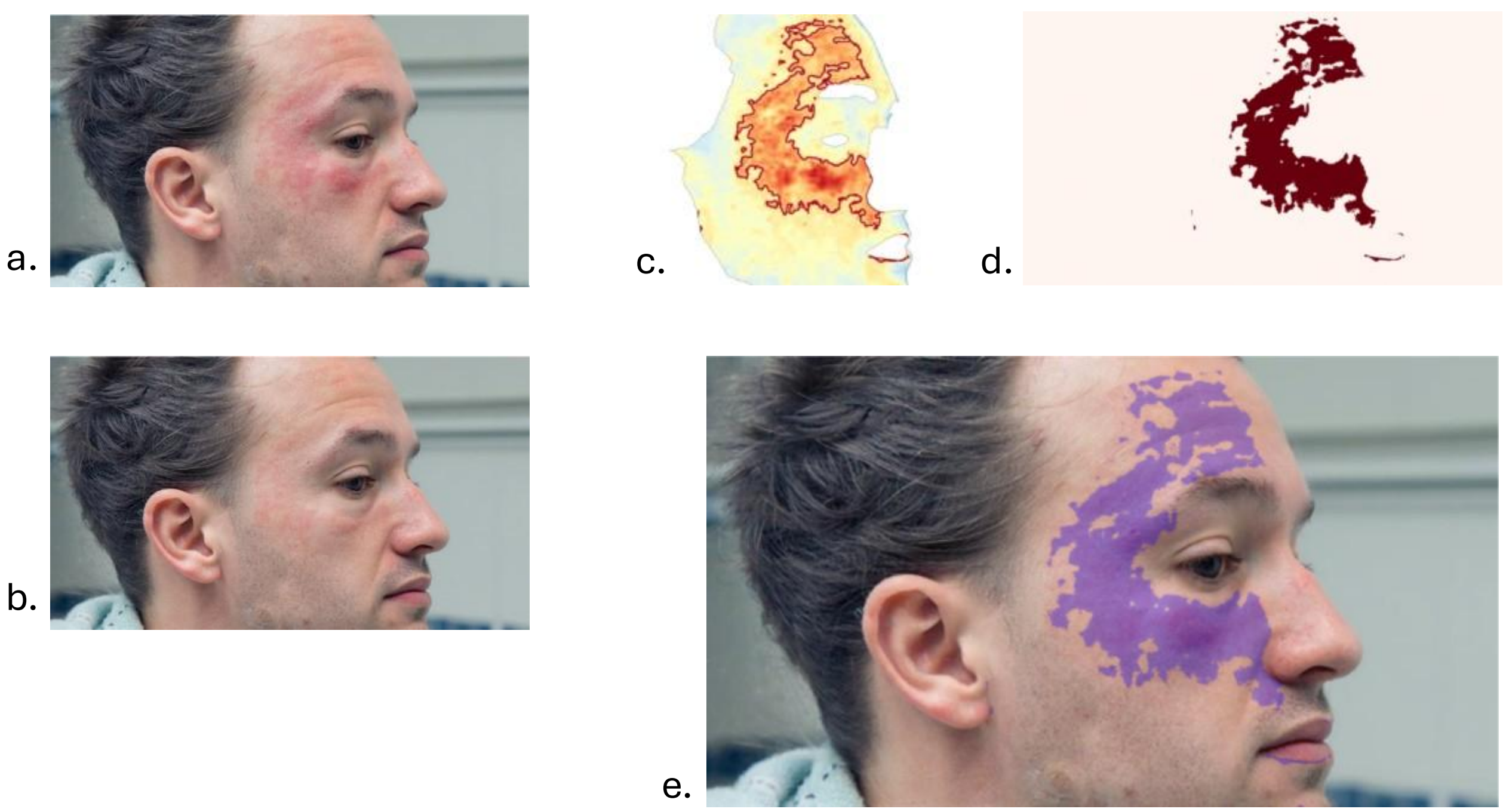

Figure 2: Overview of the Counterfactual Segment-by-Synthesis Workflow. The pipeline isolates the pathological signal by comparing the de-identified subject with a generated "healthy" version of themselves. (a) Synthetic Pathological Twin (I_path): The de-identified patient surrogate generated via FlowEdit, retaining the original erythema structure. (b) Synthetic Healthy Twin (I_healthy): A counterfactual "healthy" variant generated from the identical latent seed, ensuring pixel-perfect anatomical alignment with (a). (c) A-Channel Difference Map: The pixel-wise residual intensity calculated in the CIELAB color space (specifically the $\alpha$* channel), which effectively isolates the redness signal from skin pigmentation and lighting. (d) Extracted Mask: The final binary segmentation mask M obtained after applying the optimized dynamic threshold (θ*). (e) Annotated Surrogate: The resulting high-precision erythema mask overlaid on the synthetic twin, serving as the privacy-preserving ground truth for Federated Learning.

### 3.2. Counterfactual Segment-by-Synthesis

To generate robust training labels for the FL model, we implement the Segment-by-Synthesis paradigm [6]. For each de-identified image, the client generates a counterfactual healthy twin using the same latent seed but a modified text prompt (“healthy skin”). This yields a pair: $(I_{path},\ I_{healthy})$, as illustrated in Figs 2a and 2b. $I_{path}$ is the synthetic pathological twin resulting image, showing the operation: New Identity + Erythema. $I_{healthy}$ is the synthetic healthy twin, showing the operation: New Identity + Healthy Skin. Since identity and lighting are controlled by the fixed seed, the only variance between the two images is the pathology itself:

$$I_{path} = \mathcal{G}(z_{de-id}, \tau_{path}) \quad (2)$$

$$I_{healthy} = \mathcal{G}(z_{de-id}, \tau_{healthy}) \quad (3)$$

The Generator ($\mathcal{G}$): represents the generative model. In our case, the rectified flow transformer. This function takes a latent noise vector and a text condition as input and outputs a pixel-space image. The de-identified latent code ($z_{de-id}$) is the shared latent anchor, and this is the specific noise vector (seed) that corresponds to the synthetic surrogate identity we created in the previous step using FlowEdit. This variable appears in both equations. Because the latent code $z_{de-id}$ is held constant, the global facial geometry (nose shape, eye distance, lighting, background) remains identical in both output images. This "locks" the anatomy.

Regarding the conditioning prompts, the pathological prompt ($\tau_{path}$) is the text embedding that describes the clinical condition (e.g., "a photo of a face with skin erythema"). The counterfactual prompt ($\tau_{healthy}$) is the text embedding that describes the healthy state (e.g., "a photo of a face with clear, healthy skin"). This instructs the generator $\mathcal{G}$ to "heal" the skin while keeping everything else the same.

By defining the images $I_{path}, I_{healthy}$ this way, we ensure that the pathology itself is the mathematical difference between the two outputs:

$$I_{path} - I_{healthy} \approx \text{Pure Pathology} \quad (4)$$

Because $\mathcal{G}$ and $z_{de-id}$ are fixed, all anatomical features (which are "noise" for our segmentation task) cancel out during subtraction.

### 3.3. Differential mask extraction and calibration

The final training target is a binary mask $M$, derived via pixel-wise subtraction:

$$M = \{p \in \Omega \mid \| I_{path}(p) - I_{healthy}(p) \| > \theta\} \quad (5)$$

Here, Omega ($\Omega$) represents the entire spatial grid of the image (height × width), and $p$ represents a single pixel coordinate (x,y) within that grid. Theta ($\theta$) is the dynamic threshold of sensitivity.

This differential approach effectively filters out

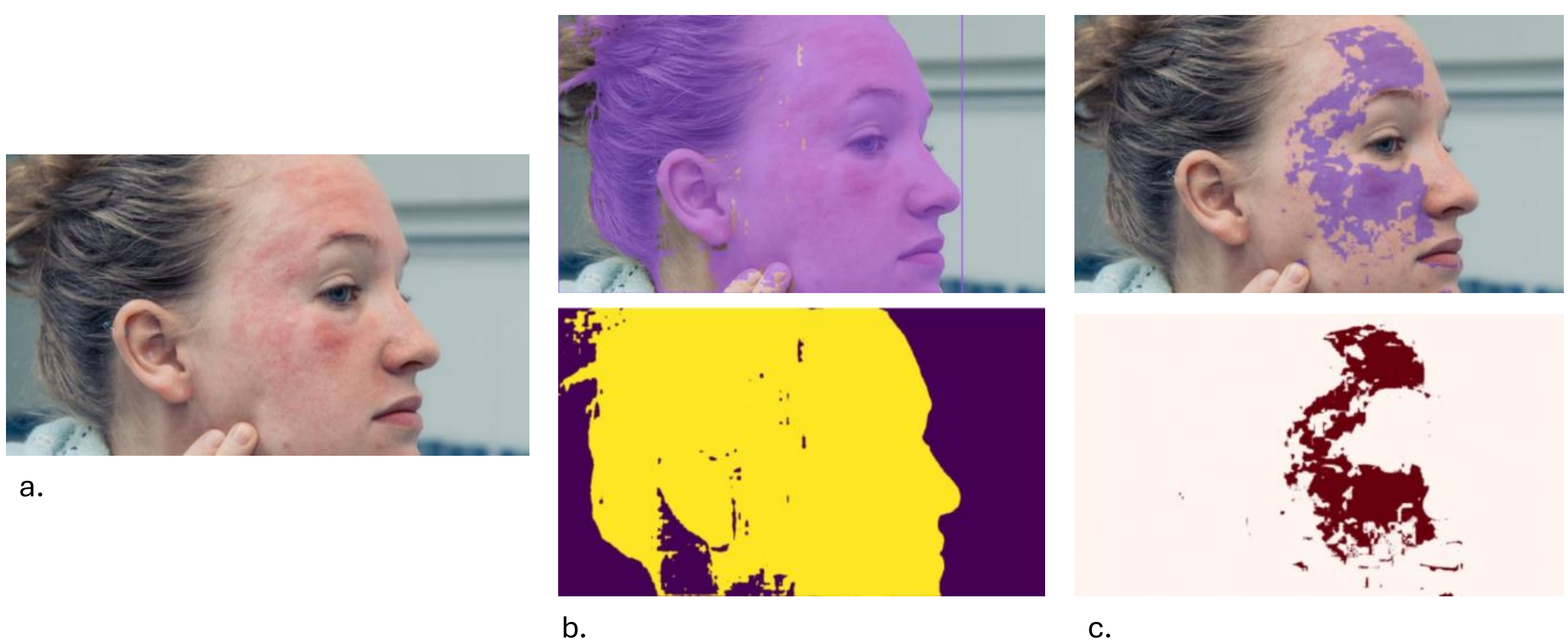

Figure 3: Qualitative comparison on diffuse facial pathology. We benchmark the robustness of our framework against the state-of-the-art Grounded HQ-SAM pipeline on a full-face clinical sample. (a) Original Patient Input ($I_{orig}$): A high-resolution profile featuring diffuse facial erythema [19]. The pathological signal is subtle and texturally similar to natural skin flushing, posing a challenge for zero-shot segmentation. (b) Baseline (Grounded HQ-SAM): The zero-shot pipeline output. Lacking a comparative “healthy” reference, the model struggles to define the amorphous boundaries of the inflammation, erroneously including non-pathological anatomical features (e.g., lips, hair) or over-segmenting onto healthy skin. (c) Segment-by-Synthesis: Our synthesis-bases privacy-preserving annotation. By subtracting the counterfactual healthy twin, the differential approach cancels out the underlying facial geometry (identity), successfully isolating the pure pathological signal without semantic confusion.

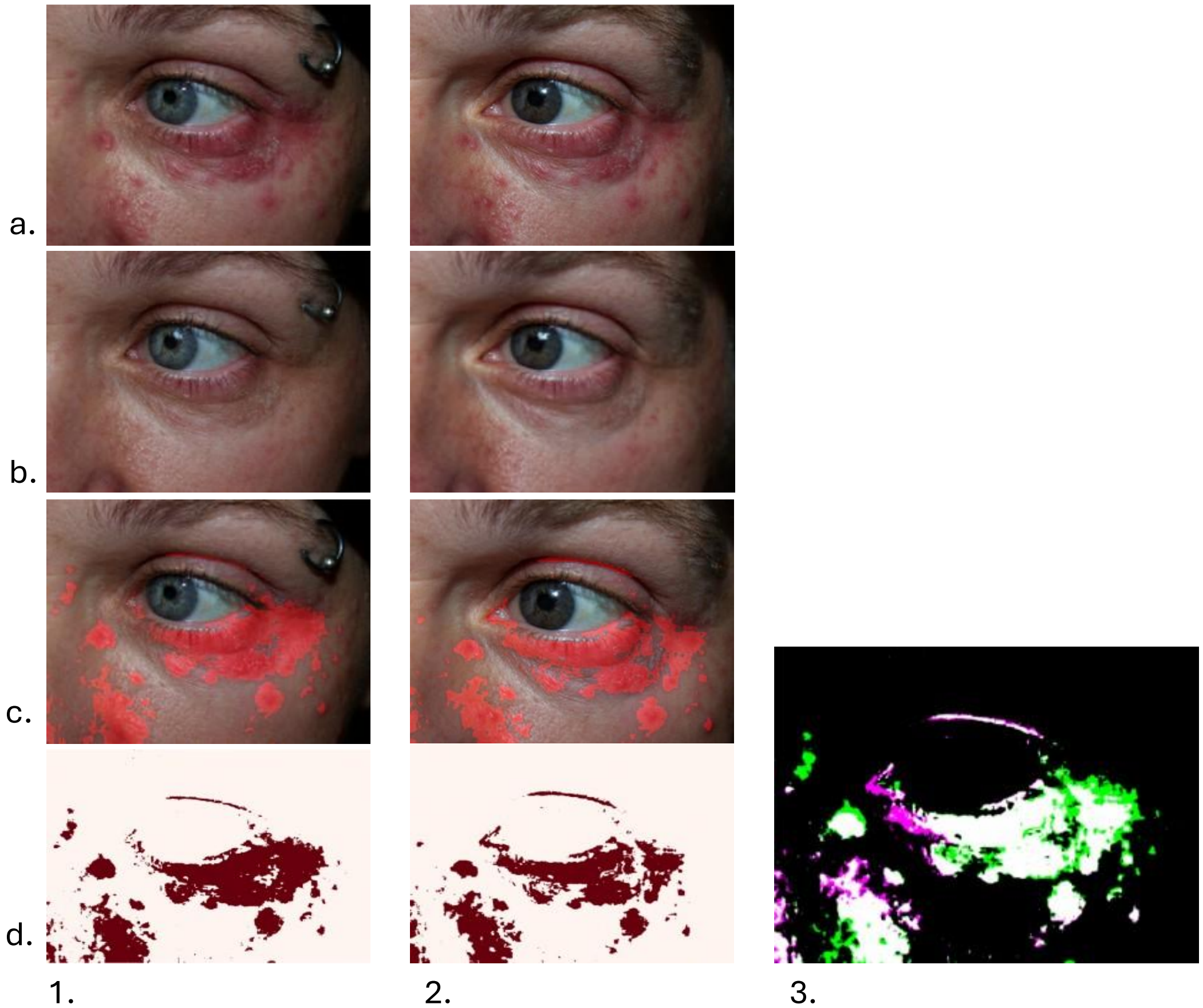

Figure 4: Validation of pathological invariance via mask stability analysis. We demonstrate that the erythema morphology remains consistent even after the identity transformation. Column 1 (Original Identity): The baseline pipeline applied to the patient. (a) Input patient image [17]. (b) Counterfactual healthy twin generated via FlowEdit. (c) Pathological annotation (mask overlay). (d) Binary segmentation mask ($M_{orig}$). Column 2 (Synthetic Surrogate): The same pipeline applied to the de-identified surrogate. (a) Transformed synthetic identity. (b) Its corresponding healthy twin. (c) Pathological annotation. (d) Binary segmentation mask ($M_{syn}$). Column 3 (Comparative Overlay): A composite visualization where the Original Mask ($M_{orig}$) is mapped to Green and the Synthetic Mask ($M_{syn}$) to Magenta. White Regions: Indicate the Intersection ($M_{orig} \cap M_{syn}$), representing perfect agreement. Green/Magenta fringes: represent morphological deviations (shift). The dominance of white pixels confirms high structural alignment (IoU$>0.67$), proving that the pathological boundaries are preserved despite the global identity shift.

"identity noise" (e.g., specific eye shapes and piercings) that typically confuses general-purpose segmenters. In a federated setting, only this De-identified Image ($I_{de-id}$) and the Synthetic Mask ($M$) are used for local training, ensuring that the original patient biometric data do not influence the model weights.

A dynamic threshold optimization is implemented to refine the segmentation accuracy. The threshold $\theta$ is optimized to match the original patient's erythema mask $M_{orig}$ as closely as possible. The optimal threshold $\theta^*$ is selected by maximizing the Intersection over Union (IoU) across the de-identified cohort:

$$\theta^* = \operatorname*{argmax}_{\theta} \frac{|M_{orig} \cap M_{de-id}(\theta)|}{|M_{orig} \cup M_{de-id}(\theta)|} \quad (5)$$

## 4. Results and Discussion: Pilot Validation

We conducted a pilot validation using high-resolution clinical samples representing challenging edge cases (e.g., facial erythema with occlusion) to assess the feasibility of this pipeline for federated learning.

### 4.1. De-identification Fidelity

The fidelity of the de-identification process is evaluated by the model's capacity to induce a significant semantic shift in the patient's identity while maintaining the structural consistency of the non-identifiable regions. This transformation is not merely a localized obfuscation but a global geometric and textural realignment of the facial architecture.

A primary benchmark for this fidelity is the transition across major biometric categories, such as the Female-to-Male identity shift illustrated in Fig. 1, while retaining the structural integrity of the erythema. This process is facilitated by the FlowEdit architecture, which allows for the modification of global identity semantics—including gender-specific jawlines and eye shapes—while anchoring local pathological textures in the latent space. The inversion-free nature of FlowEdit allowed for inference times of <20 seconds for image resolution 1024x768 (NVIDIA L4), demonstrating the potential for near real-time processing on edge nodes.

Further validation in the periocular region (Case 2) highlights the framework's ability to suppress highly specific biometric identifiers. As shown in Figure 4, the generative process successfully removes personal artifacts such as orbital piercings and specific iris patterns, while maintaining the precise morphology of the surrounding erythema. This suppression is achieved through the mixed guidance mechanism of the velocity field ($v$), where the target guidance scale ($\gamma_{tgt}$) pushes the latent geometry toward a new facial structure, effectively stripping away identifying non-pathological features.

## 4.2. Comparison with the baseline (Grounded-SAM)

We benchmarked our pipeline against a Grounded-SAM pipeline integrated with HQ-SAM [17], with qualitative comparisons shown in Fig. 3. By using HQ-SAM, we ensured that the baseline failure was not due to mask coarseness, but rather the fundamental inability of zero-shot models to distinguish pathology from semantic noise.

Failure Mode: Grounded-SAM frequently misclassifies nonpathological features (lips, piercings, and shadows) as lesions, creating "noisy labels" that degrade FL convergence.

Robustness to artifacts: The Segment-by-Synthesis approach successfully filtered out these distractors. By subtracting the "Healthy Twin," our method isolated the erythema with pixel-perfect precision, providing clean ground-truth labels for unsupervised learning.

## 4.3. Mask Stability (IoU Analysis)

We measured the Intersection over Union (IoU) between the original patient's pathology and the masks generated from the synthetic surrogates:

Facial Image [19] (Case 1): The framework achieved an average IoU of $0.685 \pm 0.002$ across diverse synthetic identities. The pathological morphology remains visually consistent despite significant changes in the underlying facial geometry (Fig. 2). This stability confirms that the "pathological signal" is invariant to the identity shift, ensuring that FL models trained on these data will generalize to real patients.

Periocular/periorificial Image [20] (Case 2): The method maintained an IoU of 0.674 (Fig. 4), even with high-contrast distractors (piercings), proving robustness against the type of "non-IID" noise common in diverse clinical datasets.

The statistical consistency of the framework was validated by comparing the raw $\alpha^*$ channel distributions across the counterfactual pipelines (Fig. 5). By analyzing the normalized histograms of the skin tones, we quantified the precision of pathology isolation and the stability of the synthetic surrogates. Distribution comparison metrics indicate that the core clinical signals remain robust, even when transitioning from real-world patient data to fully synthetic domains.

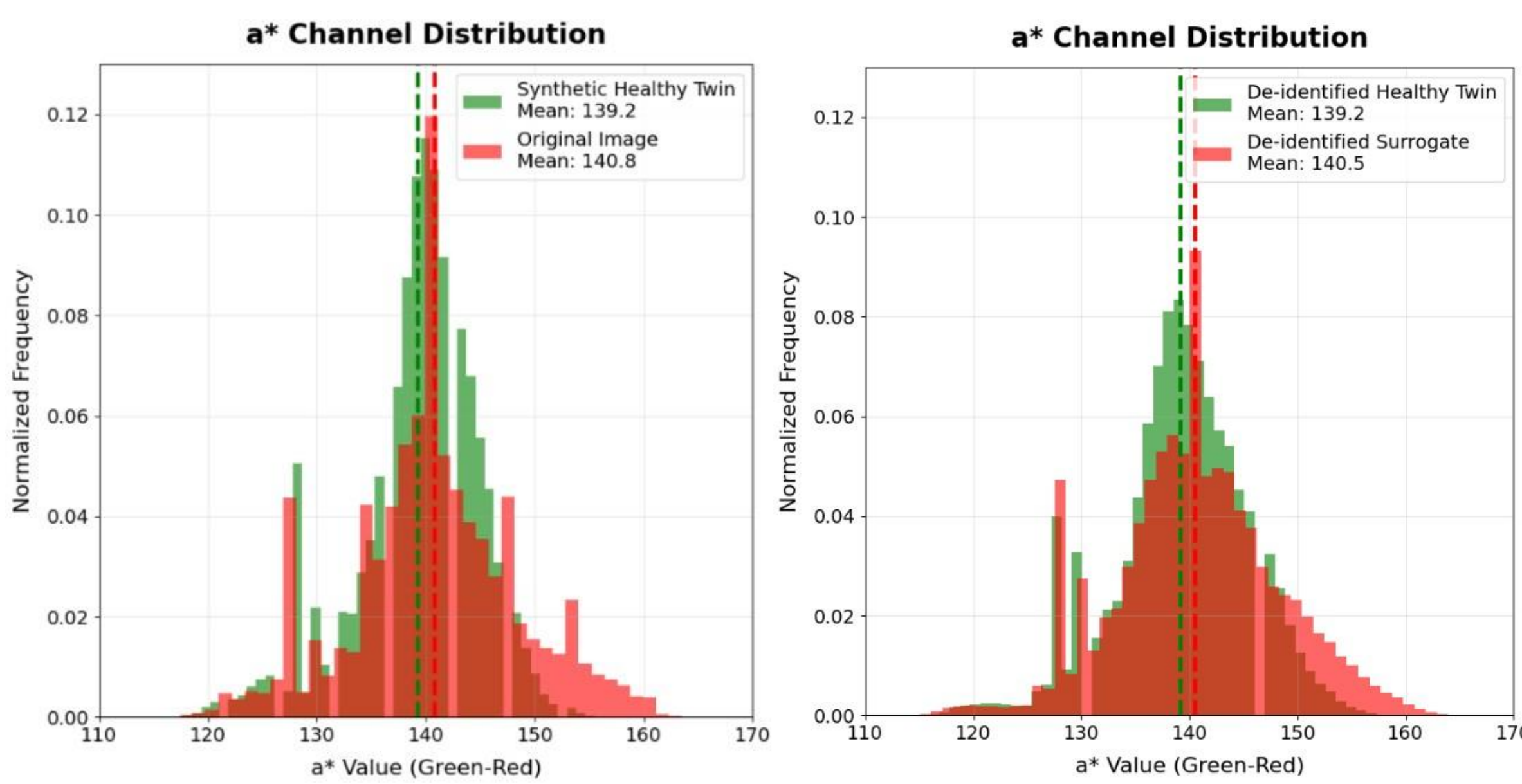

Figure 5: Normalized histograms of $\alpha$* channel intensity (Periocular Case): (Left) Original patient vs. synthetic healthy version: the high peak frequency (≈0.12) indicates that the inversion-free transformation preserves the specific statistical profile of the patient's original skin tone. (Right) synthetic pathological vs. synthetic healthy twin: The broader distribution (peak <0.10) reflects the stochastic variance introduced by the dual generative paths, despite sharing a common latent identity.

In the initial validation stage (Fig. 5, Left), the distribution of the synthetic healthy version (Fig. 4.1b) is compared directly to the original patient photograph (Fig 4.1a). The high peak density of 0.12 and a standard deviation of 5.69 indicate that the inversion-free transformation successfully performs localized "healing". By maintaining most of pixels within the patient's original color gamut (mean $\alpha^* \approx 139$), the model can isolate erythema without distorting the underlying healthy skin texture or lighting of the specific patient.

The comparison between the two synthetic twins (Fig. 5, Right, corresponding to Figs 4.2a and 4.2b), exhibits a broader variance and a lower peak density (<0.10). This phenomenon is attributed to the "Double Rendering" effect; the pixel-wise stochasticity does not perfectly overlap because both twins are fully synthetic outputs of separate FlowEdit traversals. However, the higher Bhattacharyya coefficient (0.977 vs. 0.959) and lower KS statistic (0.103 vs. 0.130) for this synthetic pair indicate that the two generative outputs are statistically more compatible with each other than the synthetic surrogate is with the original photograph. This confirms that the "Digital Twin" approach provides a highly stable baseline for extracting differential masks that are free from the biometric interference of the original patient data.

## 4.4. Qualitative Analysis: Viewing Angles and Feature Persistence

Our pilot analysis reveals that the efficacy of prompt-guided de-identification is sensitive to the clinical viewing angle. We observed a distinct performance divergence between the Facial Image [19] (Case 1) and the Periocular/periorificial Image [20] (Case 2).

The Macro-Lens Challenge: In the close-up regime (Case 2), prompt engineering and hyperparameter tuning proved to be more difficult. The generative model struggled to anchor the semantic context due to the lack of global facial landmarks (e.g., chin contour, hair, background depth). Without these cues, the model occasionally failed to disentangle the "subject" from the "background," a limitation consistent with previous zero-shot Segment-by-Synthesis observations [6].

Biometric Feature Persistence: We also identified the limitations of fine-grained feature control during single-pass editing. In the full-face sample (Case 1), we explicitly attempted to alter ear morphology—a robust biometric soft-identifier—via prompt engineering. Despite these attempts, the ear structure remained largely unchanged in the synthetic surrogate. The same was observed for the eyebrow morphology in the periocular/periorificial image. This indicates that a single velocity field may be insufficient for total biometric suppression.

A potential solution is a multistage de-identification pipeline, where intermediate editing steps specifically target resistant features (e.g., ears or hairline) before the final pathological consistency check is applied.

## 4.5. Broader Clinical Generalizability

Although this study primarily focuses on erythema, the proposed Segment-by-Synthesis de-identification methodology is inherently pathology-agnostic. The paradigm can be extended to any clinical condition characterized by visual alterations during disease progression that can be captured via high-resolution clinical imaging.

As highlighted in recent systematic literature reviews [2], the quantitative assessment and automated detection of conditions such as dermatitis, vitiligo, and alopecia areata remain a significant bottleneck in medical informatics. Our framework addresses this gap by enabling the generation of privacy-compliant "Digital Twins" that decouple the pathological signal from the biometric identity of the patient.

The proposed framework is readily extensible to other dermatological conditions, such as:

Pigmentation Disorders: The counterfactual healthy twin logic can be used to restore melanocyte-rich skin regions in vitiligo, allowing for precise quantification of depigmented patches.

Inflammatory Conditions: Hair-loss density and follicular inflammation patterns in alopecia areata can be isolated by comparing the pathological surrogate with a reconstructed healthy scalp baseline.

Skin Monitoring: The pipeline could be adapted to clinical images to monitor lesions growth, border irregularity, and morphological changes in melanoma, ensuring that sensitive biometric data are never exposed during the multi-institutional training required for rare cancer detection.

By preserving fine-grained diagnostic markers—including psoriatic scaling, eczematous texture, and pigment distribution—while stochastically transforming global biometric geometry, the pipeline effectively decouples pathology from identity, facilitating high-fidelity collaborative research and multi-center clinical data analysis. This modularity is particularly vital for rare diseases where privacy constraints often lead to severe data scarcity, as it allows for the creation of standardized, de-identified synthetic datasets that retain full diagnostic utility.

## 5. Conclusion

This study introduced a novel, inversion-free generative framework designed to address the fundamental conflict between patient privacy and diagnostic utility in clinical dermatological imaging. By leveraging Rectified Flow Transformers, we demonstrated that high-fidelity de-identification can be achieved with significantly lower computational latency than traditional diffusion-based inversion methods, making the pipeline suitable for real-time deployment at the clinical edge.

The introduction of the Segment-by-Synthesis mechanism represents a significant advancement in unsupervised medical image annotation. By generating counterfactual 'healthy' and 'pathological' digital twins, the framework successfully decouples sensitive biometric signals from critical diagnostic markers, achieving high pathological preservation with an Intersection over Union (IoU) stability exceeding 0.67. Our results indicate that this approach not only ensures rigorous privacy compliance—aligning with global standards such as GDPR—but also maintains the structural and textural integrity of pathological signals.

The implications of this work extend beyond identity protection; the proposed pipeline provides a scalable solution for the creation of high-quality, de-identified synthetic datasets. Such datasets are vital for training robust computer-aided diagnosis (CAD) systems, especially for rare dermatological conditions where data scarcity is often exacerbated by privacy restrictions. Future research will focus on validating the framework across a broader spectrum of skin pathologies, including melanoma and chronic inflammatory disorders, and evaluating its performance in diverse multi-institutional collaborative environments. Ultimately, this generative de-identification paradigm paves the way for a more secure and transparent era of biomedical signal processing, where the sharing of clinical insights no longer necessitates the exposure of patient identity